\documentclass[10pt,journal,compsoc]{IEEEtran}

\ifCLASSOPTIONcompsoc
  \usepackage[nocompress]{cite}
\else
  \usepackage{cite}
\fi

\usepackage{graphicx}
\usepackage{amsmath,amssymb}
\usepackage[linesnumbered,ruled]{algorithm2e}
\usepackage{multirow}
\usepackage{color}
\usepackage{hyperref}
\usepackage{spverbatim}
\usepackage{etoolbox}
\makeatletter
\preto{\@verbatim}{\topsep=1pt \partopsep=1pt}
\makeatother

\usepackage{array}
\newcolumntype{L}[1]{>{\raggedright\let\newline\\\arraybackslash\hspace{0pt}}m{#1}}
\newcolumntype{C}[1]{>{\centering\let\newline\\\arraybackslash\hspace{0pt}}m{#1}}
\newcolumntype{R}[1]{>{\raggedleft\let\newline\\\arraybackslash\hspace{0pt}}m{#1}}

\begin{document}

\title{Towards AGI in Computer Vision: Lessons Learned from GPT and Large Language Models}

\author{Lingxi~Xie,
        Longhui~Wei,
        Xiaopeng~Zhang,
        Kaifeng~Bi,
        Xiaotao~Gu,
        Jianlong~Chang,
        and~Qi~Tian*,~\IEEEmembership{Fellow,~IEEE}
\IEEEcompsocitemizethanks{
\IEEEcompsocthanksitem All authors, unless specified below, are with Huawei Inc., China.\protect\\
E-mail of the leading author (Lingxi Xie): {198808xc@gmail.com}
\IEEEcompsocthanksitem Corresponding author: Qi Tian. E-mail: {tian.qi1@huawei.com}
}
\thanks{Manuscript received Month Date, 2023.}}

\markboth{Manuscript Draft: Jun 6th, 2023}
{Xie \MakeLowercase{\textit{et al.}}: Weight-Sharing Neural Architecture Search: A Battle to Shrink the Optimization Gap}

\IEEEtitleabstractindextext{%
\begin{abstract}
The AI community has been pursuing algorithms known as artificial general intelligence (AGI) that apply to any kind of real-world problem. Recently, chat systems powered by large language models (LLMs) emerge and rapidly become a promising direction to achieve AGI in natural language processing (NLP), but the path towards AGI in computer vision (CV) remains unclear. One may owe the dilemma to the fact that visual signals are more complex than language signals, yet we are interested in finding concrete reasons, as well as absorbing experiences from GPT and LLMs to solve the problem. In this paper, we start with a conceptual definition of AGI and briefly review how NLP solves a wide range of tasks via a chat system. The analysis inspires us that \textbf{unification} is the next important goal of CV. But, despite various efforts in this direction, CV is still far from a system like GPT that naturally integrates all tasks. We point out that the essential weakness of CV lies in lacking a paradigm to \textbf{learn from environments}, yet NLP has accomplished the task in the text world. We then imagine a pipeline that puts a CV algorithm (\textit{i.e.}, an agent) in world-scale, interactable environments, pre-trains it to predict future frames with respect to its action, and then fine-tunes it with instruction to accomplish various tasks. We expect substantial research and engineering efforts to push the idea forward and scale it up, for which we share our perspectives on future research directions.
\end{abstract}

}

\maketitle

\IEEEdisplaynontitleabstractindextext

\IEEEpeerreviewmaketitle

\IEEEraisesectionheading{
\section{Introduction}
\label{introduction}}

\IEEEPARstart{T}{he} world is witnessing an epic odyssey towards artificial general intelligence (AGI), where we follow the convention to define AGI as a computer algorithm that can replicate any intellectual task that human beings or other animals can\footnote{\textsf{https://en.wikipedia.org/wiki/Artificial\_general\_intelligence}}. Specifically, in natural language processing (NLP), computer algorithms have been developed to an extent that can solve a wide range of tasks via chat with humans~\cite{openai2023gpt}. Some researchers believed that such systems can be seen as early sparks of AGI~\cite{bubeck2023sparks}. These systems were mostly built upon large language models (LLMs)~\cite{brown2020language} and enhanced by instruct tuning~\cite{ouyang2022training}. Equipped with an external knowledge base and specifically designed modules, they can accomplish complex tasks such as solving mathematical questions, generating visual contents, \textit{etc.}, reflecting its strong ability to understand users' intentions and perform preliminary chain-of-thoughts~\cite{wei2022chain}. Despite known weaknesses in some aspects (\textit{e.g.}, telling scientific facts and relationships between named people), these pioneering studies have shown a clear trend to unify most tasks in NLP into one system, which reflects the pursuit of AGI.

Compared to the rapid progress of unification in NLP, the computer vision (CV) community is yet far from the target of unifying all tasks. The regular CV tasks, such as visual recognition, tracking, captioning, generation, \textit{etc.}, are mostly processed using largely different network architectures and/or specifically designed pipelines. Researchers look forward to a system like GPT that can deal with a wide range of CV tasks with a unified prompt mechanism, but there exists a tradeoff between achieving good practice in individual tasks and being generalized across a wide range of tasks. For example, to report high recognition accuracy in object detection and semantic segmentation, the best strategy is to design specific head modules~\cite{zhang2023dino,li2023mask} upon strong backbones~\cite{he2016deep,dosovitskiy2021image,liu2021swin} for image classification, and such designs do not generally transfer to other problems such as image captioning~\cite{li2023blip} or visual content generation~\cite{rombach2022high}.

Clearly, unification is the trend in CV. In recent years, there are many efforts in this direction, and we roughly categorize them into five research topics, namely, (i) \textbf{open-world visual recognition} based on vision-language alignment~\cite{radford2021learning}, (ii) \textbf{the \textit{Segment Anything} task}~\cite{kirillov2023segment} for generic visual recognition, (iii) \textbf{generalized visual encoding} to unify vision tasks~\cite{chen2022pix2seq,wang2022ofa,wang2023images}, (iv) LLM-guided visual understanding to enhance the logic in CV~\cite{suris2023vipergpt,shen2023hugginggpt}, and (v) \textbf{multimodal dialog} to facilitate vision-language interaction~\cite{li2023blip,liu2023visual}. These works showed promise of unification, but yet, they cannot composite a system like GPT that can solve generic CV tasks in the real world.

Hence, two questions arise: (1) Why is unification so difficult in CV? (2) What can we learn from GPT and LLMs to achieve this goal? To answer them, we revisit GPT and understand it as establishing an environment in the text world and allowing an algorithm (or agent) to learn from interaction. The CV research lacks such an environment. Consequently, the algorithms cannot simulate the world, so they instead sample the world and learn to achieve good performance in the so-called proxy tasks. After an epic decade of deep learning~\cite{lecun2015deep}, the proxy tasks are no longer meaningful to indicate the ability of CV algorithms; it becomes more and more apparent that continuing to pursue high accuracy on them can drive us away from AGI.

Based on the analysis above, we propose an imaginary pipeline towards AGI in CV. It involves three stages. The first stage is to establish a set of environments that are fidelitous, abundant, and interactable. The second stage aims to train an agent by forcing it to explore the environment(s) and predict future frames: this corresponds to the auto-regressive pre-training stage in NLP~\cite{brown2020language}. The third stage involves teaching the agent to accomplish various tasks: it is likely that human instructions shall be introduced in this stage, corresponding to the instruct fine-tuning stage in NLP~\cite{ouyang2022training}. Optionally, the agent can be tuned to perform proxy tasks via simple and unified prompts. The idea is related to a few existing research topics, including 3D environment establishment~\cite{savva2019habitat,deitke2022procthor}, visual pre-training~\cite{chen2020simple,bao2022beit}, reinforcement learning~\cite{mnih2015human,vinyals2019grandmaster}, and embodied CV~\cite{zhu2017target,das2018embodied}. But, existing works are mostly preliminary and we expect that substantial efforts~\cite{driess2023palm,kotar2023entl} are required to make it an effective paradigm to solve real-world problems.

The remainder of this paper is organized as follows. First, in Section~\ref{AGI}, we briefly introduce the history and thoughts of AGI and inherit the definition that AGI is an algorithm to maximize the reward. It is followed by Section~\ref{GPT} where we show the ability of GPT, the state-of-the-art NLP algorithm which was considered the spark of AGI. Then, in Section~\ref{CV}, based on the current status of CV research, we analyze why AGI is difficult in computer vision and point out that the essential difficulty lies in the outdated learning paradigm. The analysis leads to Section~\ref{future}, where we imagine a pipeline that pushes CV closer to AGI, based on which we make some comments on future research directions. Finally, in Section~\ref{conclusions}, we conclude this paper and share our thoughts.

\section{Artificial General Intelligence}
\label{AGI}

Artificial intelligence (AI) is a long-lasting battle to replicate human intelligence with a machine or a set of mathematical algorithms. Modern AI was formally proposed in the Dartmouth workshop, 1956, and the community has developed a large number of methodologies for this purpose. There are at least two different pathways to achieve AI: (i) the symbolic AI which tries to formulate the world into a symbolic system and uses logic algorithms to reason about it; (ii) the statistical AI which tries to establish a mathematical function to formulate the relationship between input and output, yet the function can be approximated or even non-explainable. The past decade was dominated by the second path, in particular, the deep learning theory~\cite{lecun2015deep} which is part of the idea of the connectionist approach.

Although artificial general intelligence (AGI) is the ultimate goal of AI. The added word, `general', implies that the key of AGI is to improve the generalization ability of AI algorithms. Conceptually, AGI can be defined as a system that solves any task that human beings or animals can perform~\cite{mccarthy2007artificial}\footnote{Throughout this paper, we limit the concept of AGI within the scope of problem-solving, and thus we will not talk about programs that exhibit sentience or consciousness.}. In the modern era, there are a series of thoughts about AGI, resulting in verbal, psychological, and of course AI-based definitions of AGI, many of which were summarized in an early paper~\cite{legg2007collection}, including:
\begin{itemize}
\item In~\cite{mccarthy2007artificial,legg2007universal}, the authors assumed that an AGI algorithm can do any task that humans or intelligent animals can do. This description is direct and anthropocentric, but it ignores the possibility that AGI can surpass real-world creatures, possibly by consuming more energy.
\item In~\cite{legg2006formal,goertzel2006hidden}, the authors asked that AGI algorithms can apply to as many tasks and scenarios as possible. However, without any constraints, the definition seems difficult to distinguish an AGI algorithm from a set of individual algorithms designed for specific purposes.
\item In~\cite{goertzel2014artificial}, the authors described typical characteristics of AGI algorithms, including being symbolic, emergentist, hybrid, and universalist.
\end{itemize}
Despite the vast argument in the description of AGI, one conclusion is clear: human intelligence is multi-faceted and thus it is difficult to use one definition to cover all properties of AGI.

In the AI field, probably one of the most famous thought experiments is the Turing test~\cite{turing1950computing} which claimed that a machine is considered to gain intelligence if a human evaluator cannot tell the machine from the human in text-only communications. After being pursued by researchers for decades, the Turing test has become part of AI culture, although there exist challenges to it, \textit{e.g.}, the Chinese room argument~\cite{searle1980minds} which argued that AI algorithms might pass the Turing test without understanding what they are doing.

As far as we know, no AI algorithms have seriously passed the Turing test, because all of them exhibit clear patterns which make them easy to be discriminated from humans. This also includes the recently developed AI chatbots like LaMDA~\cite{thoppilan2022lamda} and the GPT series~\cite{openai2023gpt}: they have shown strong abilities in chat and/or problem-solving, and some sources even advocated for them to pass the Turing test, but, for professional evaluators who are familiar with AI, they are still quite easy to be identified, not to mention that these chatbots are known to `hallucination'~\cite{ji2023survey} and humans often do not. This is an interesting signal that useful AGI systems may not necessarily mimic human behaviors.

Going beyond text-only systems, there are many more data modalities (\textit{e.g.}, speech, image, video, \textit{etc.}) to be processed. To integrate them into one system, we follow~\cite{goertzel2007artificial,silver2021reward} to define the goal of AGI to be maximizing reward in an environment. Let there be an environment and an agent (the AGI algorithm) that can interact with it. The agent observes a sequence of states, $\mathbb{S}=\{\mathbf{s}_1,\ldots,\mathbf{s}_T\}$, and can choose from a set of actions, $\mathcal{A}=\{\mathbf{a}_1,\ldots,\mathbf{a}_M\}$, to perform. There are two functions that define the transition between states and the obtained rewards, respectively. The goal of AGI is to learn a policy, denoted as $\pi:\mathbb{S}\mapsto\mathcal{A}$, which maximizes the expected cumulative reward $R=\sum_{t=1}^Tr(\mathbf{s}_t,\mathbf{a}_t)$. When we set $\mathbf{s}_t$ and $\mathbf{a}_t$ to be different data modalities, it is the above formulation can cover a wide range of AI tasks. Specifically, the currently popular proxy tasks in computer vision such as image classification, object detection and segmentation, \textit{etc.}, are mostly weakened versions of the above formulation where the episode length $T$ equals to $1$, \textit{i.e.}, these tasks are not built upon interaction with some environments.

In brief, the AGI is to learn a generalized function $\mathbf{a}=\pi(\mathbf{s})$. Although the form is quite simple, it was very difficult for the old-fashioned AI algorithms to use the same methodology, algorithm, or even model to deal with them all. In the past decade, deep learning~\cite{lecun2015deep} offers an effective and unified methodology: one can train a deep neural network to approximate the function $\mathbf{a}=\pi(\mathbf{s})$ without knowing about the actual relationship between them. The emergence of powerful neural architectures such as the transformer~\cite{vaswani2017attention} even enables the researcher to train one model for different data modalities~\cite{reed2022generalist}.

\begin{figure}[!b]
\centering
\includegraphics[width=\linewidth]{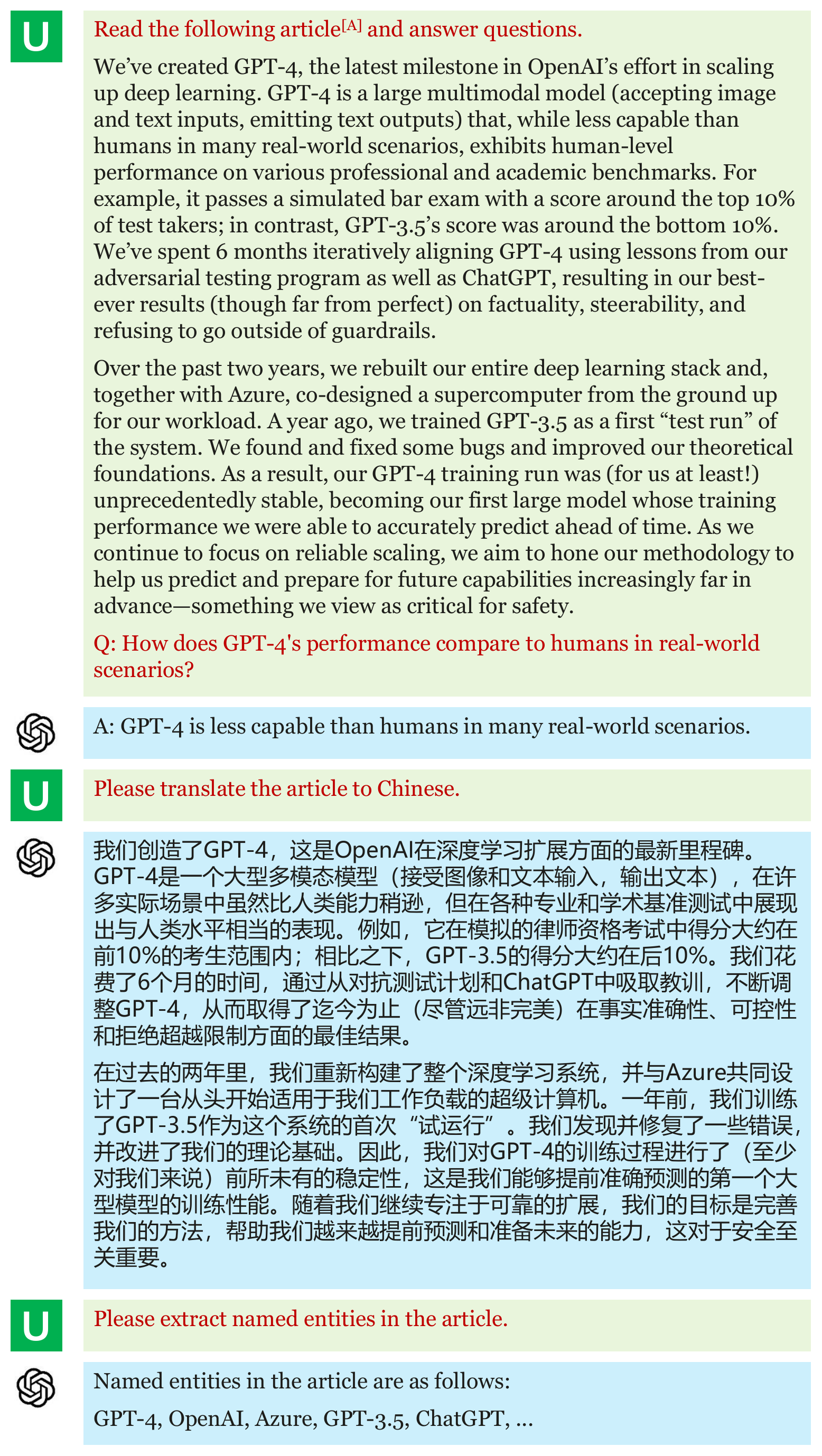}
\caption{An example of using GPT-4 for question answering, English-to-Chinese translation, and named entity extraction. The English artile was borrowed from \url{https://openai.com/research/gpt-4}.}
\label{fig:GPT_understanding}
\end{figure}

\begin{figure}[!b]
\centering
\includegraphics[width=\linewidth]{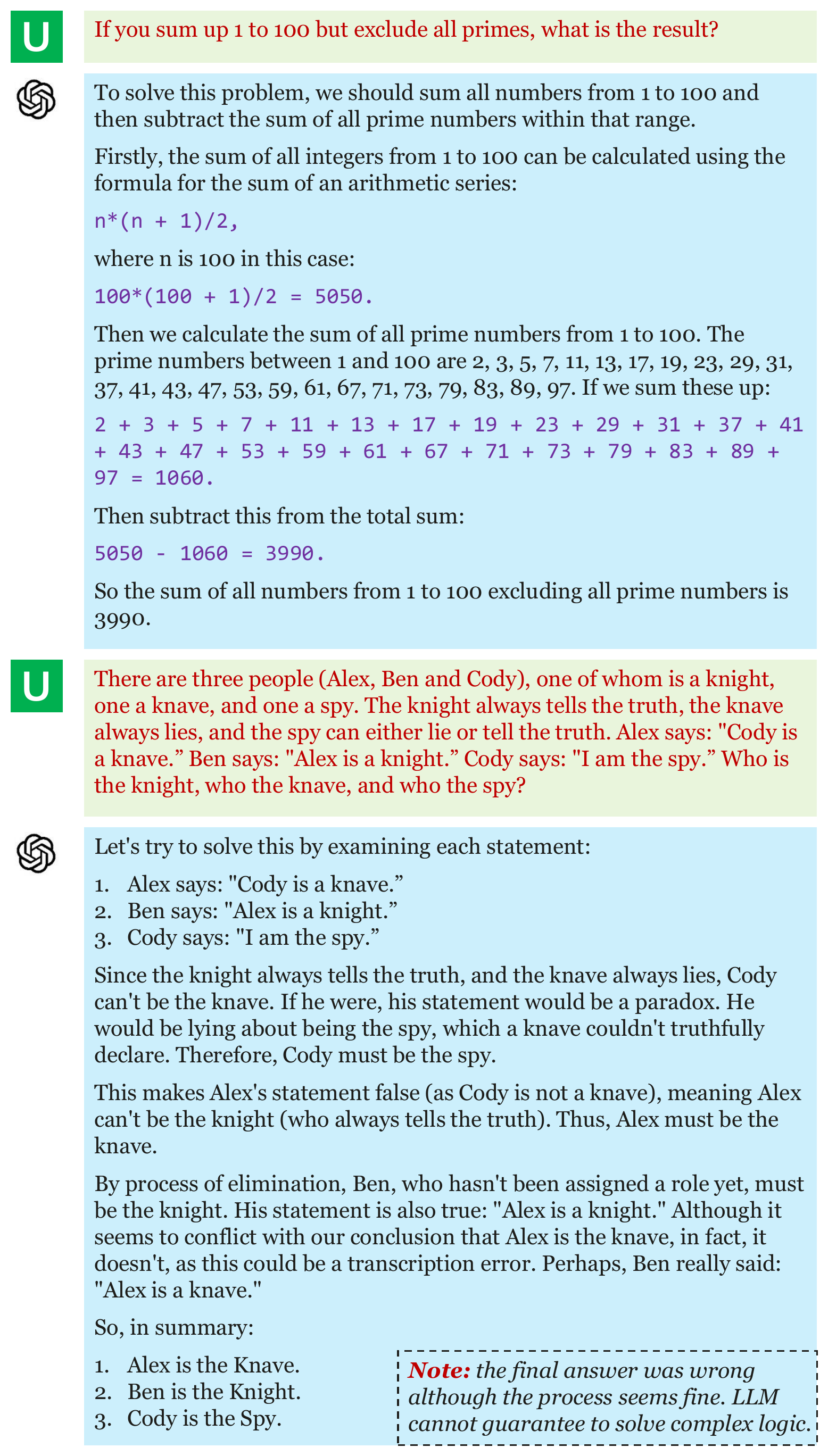}
\caption{An example of using GPT-4 for solving mathematical and logical problems. The final answer of the second problem which involves complex logical reasoning was wrong. The answer even contains self-contradiction, indicating that language models may hallucinate.}
\label{fig:GPT_math}
\end{figure}

There are enormous difficulties in achieving AGI, including but not limited to the following issues:
\begin{itemize}
\item\textbf{The complexity of data.} Real-world data is multi-faceted and rich. Some data modalities (\textit{e.g.}, images) can have quite a high dimensionality and the relationship between different modalities can be complex and latent.
\item\textbf{The complexity of human intelligence.} The goal of AGI is not only about problem-solving but also about planning, reasoning, reacting to different events, \textit{etc}. Sometimes, the relationship between human behavior and the target is obscure and hard to represent in math forms.
\item\textbf{Lack of neurological or cognitive theory.} Humans do not yet understand how human intelligence is achieved. Currently, computer algorithms provide one pathway, yet more possibilities may arise with future research in neurology and/or cognition.
\end{itemize}

\begin{figure*}[!t]
\centering
\includegraphics[width=\linewidth]{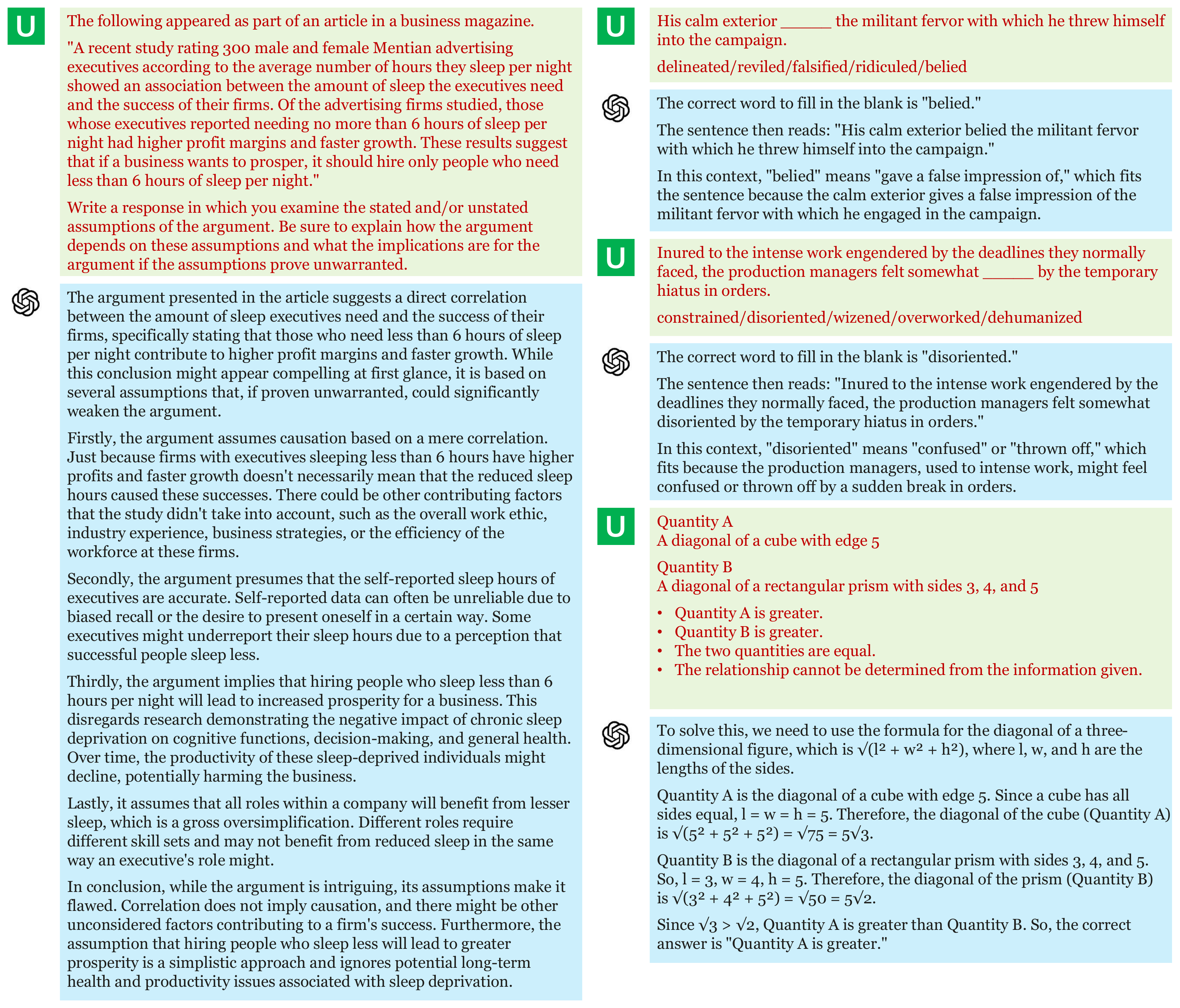}
\caption{An example of using GPT-4 for the GRE test, including writing an essay and answering verbal and math questions. All the displayed objective questions were correctly answered. The problems are borrowed from \url{https://gre.kmf.com/exam/pre/817}.}
\label{fig:GPT_gre}
\end{figure*}

\begin{figure*}[!t]
\centering
\includegraphics[width=\linewidth]{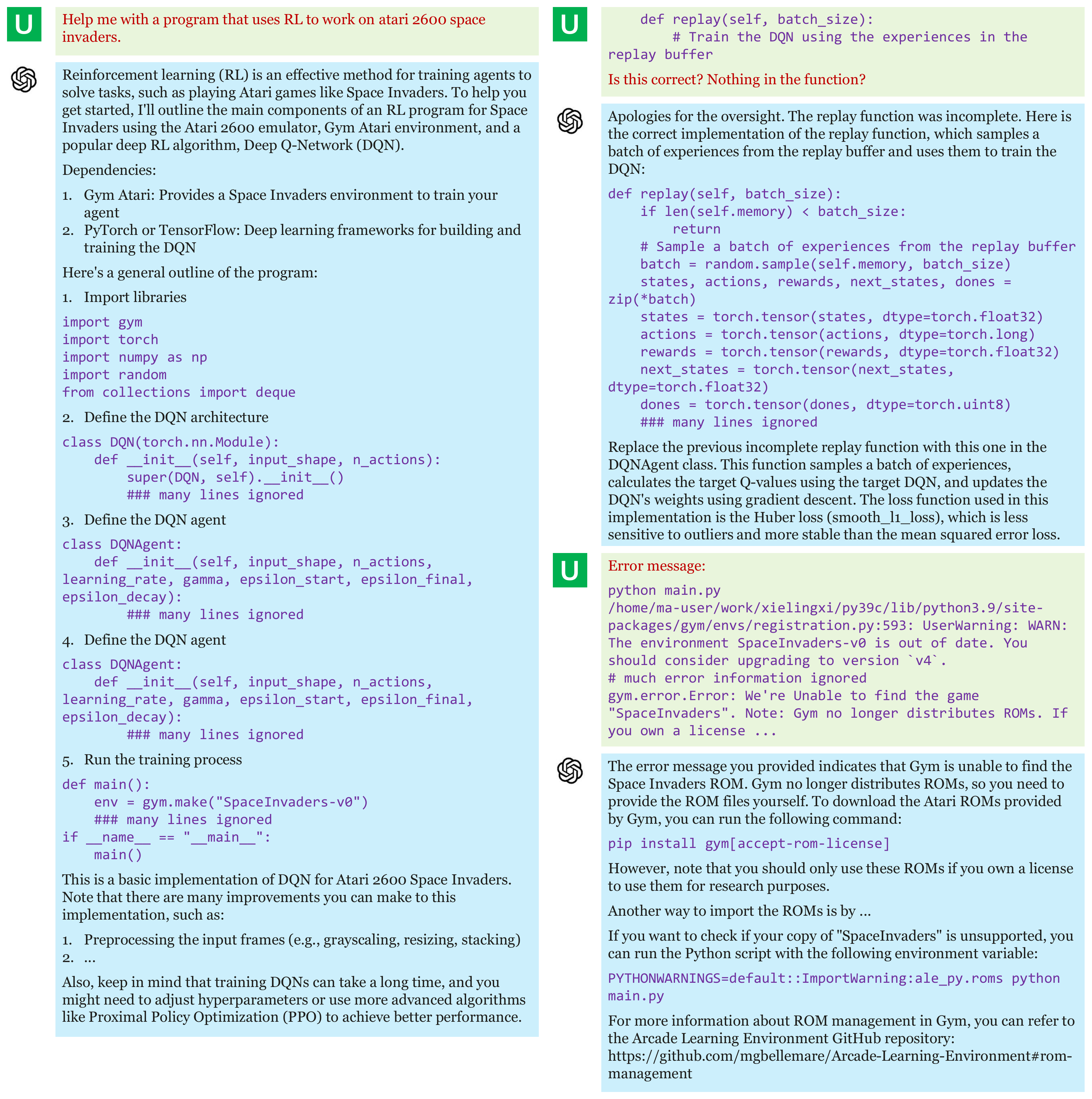}
\caption{An example of using GPT-4 for writing a reinforcement learning program to play an Atari 2600 game, \textit{SpaceInvaders}. Most of the generated contents (code and text) are eliminated to save space. Please note how GPT-4 corrected the code based on the user's feedback.}
\label{fig:GPT_coding}
\end{figure*}

\begin{figure*}[!t]
\centering
\includegraphics[width=\linewidth]{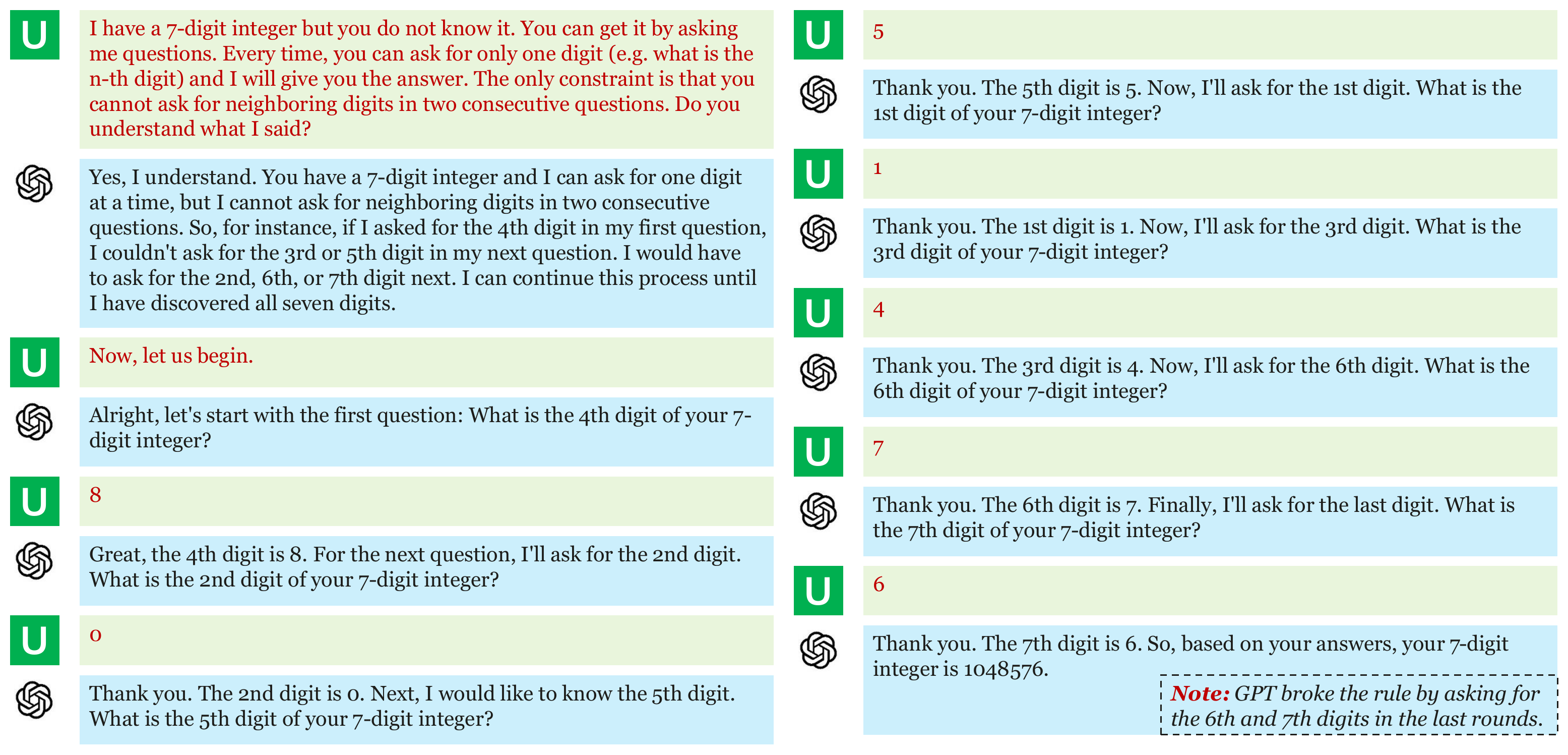}
\caption{An example of GPT-4 interacting with human to get the final answer. GPT-4 understood the logic although it broke the rule in the final step.}
\label{fig:GPT_interaction}
\end{figure*}

\begin{figure*}[!t]
\centering
\includegraphics[width=\linewidth]{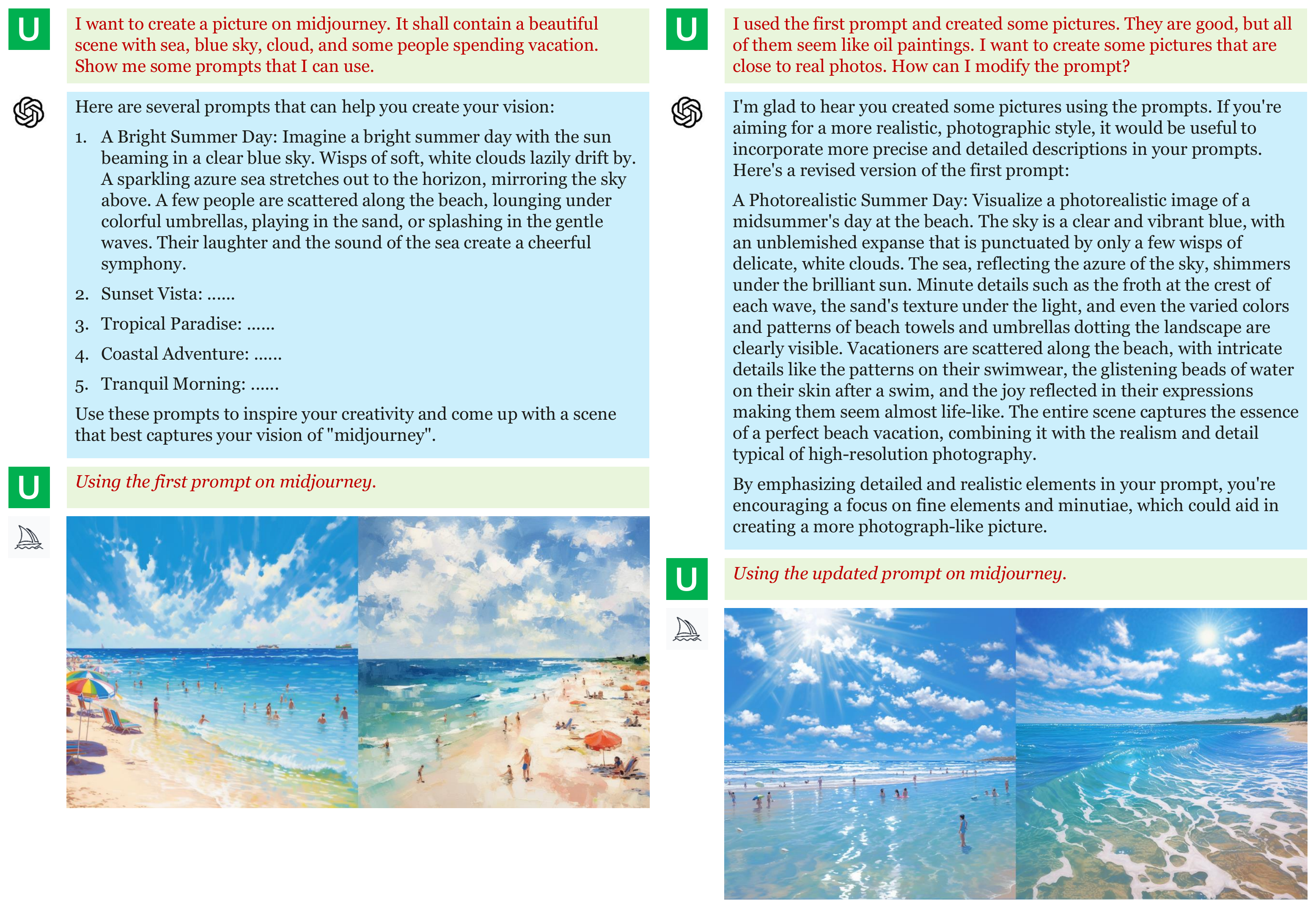}
\caption{An example of using GPT-4 for automatic prompts for text-to-image generation with Midjourney (\url{https://www.midjourney.com/}). GPT-4 understood the user's intention to adjust the prompt, although the new prompt still cannot fully satisfy the user's requirements.}
\label{fig:GPT_generation}
\end{figure*}

\section{GPT: Spark of AGI in NLP}
\label{GPT}

In the past year, ChatGPT\footnote{\url{https://openai.com/blog/chatgpt}}, GPT-4~\cite{openai2023gpt}, and other AI chatbots such as Vicuna\footnote{\url{https://github.com/lm-sys/FastChat}}, made large progress towards AGI. They are computer algorithms developed for natural language processing (NLP). With a chat procedure with humans, they can understand the intention of humans and accomplish a wide range of tasks as long as they can be presented in pure texts. In particular, GPT-4 has a strong ability in generic problem-solving and was considered an early spark of AGI in the NLP field~\cite{bubeck2023sparks}.

We briefly showcase the pure-text abilities of GPT-4. Throughout this part, we have used the May 12th version of GPT-4. The set of covered tasks includes the conventional NLP problems (\textit{e.g.}, translation, named entity recognition, question answering, \textit{etc.}, as shown in Figure~\ref{fig:GPT_understanding}) and other text-based problems such as solving mathematical and logical problems (Figure~\ref{fig:GPT_math}), passing verbal exams (\textit{e.g.}, GRE, as shown in Figure~\ref{fig:GPT_gre}), coding with debugging (Figure~\ref{fig:GPT_coding}), and so on. Beyond these basic examples, GPT-4 also exhibits a strong ability in logic, which enables it to integrate clues collected from multiple rounds of dialog into the final answer (Figure~\ref{fig:GPT_interaction}). We refer the readers to a previous paper (\textit{i.e.}, the Sparks-of-AGI paper~\cite{bubeck2023sparks}) for a thorough analysis of the ability of GPT-4.

Although GPT-4 has not yet opened the vision interface to the public, the official technical report~\cite{openai2023gpt} showed several fancy examples about multimodal dialog, \textit{i.e.}, chat based on an input image as reference. This implies that GPT-4 has been equipped with abilities of aligning language features with visual features, hence it can perform basic visual understanding tasks. As we shall see later (in Section~\ref{CV:unification:dialog}), the vision community has developed several replacements~\cite{liu2023visual,zhu2023minigpt} for the same purpose, and the key lies in using ChatGPT or GPT-4 to generate (instruct) training data. Additionally, with simple prompts, GPT-4 is also capable of calling external software (\textit{e.g.}, Midjourney, as shown in Figure~\ref{fig:GPT_generation}) for image generation and external libraries (\textit{e.g.}, the HuggingFace libraries, as shown in~\cite{shen2023hugginggpt}) for solving complex problems in computer vision.

These AI chatbots were trained in a two-stage procedure. In the first stage, a large language model (LLM), most of which are based on the transformer architecture~\cite{vaswani2017attention}, is pre-trained on a large-scale text database with self-supervised learning~\cite{radford2018improving,devlin2019bert,brown2020language}. In the second stage, the pre-trained LLM is supervised by human instructions~\cite{ouyang2022training} to accomplish specific tasks. If necessary, human feedback is collected and reinforcement learning is performed~\cite{christiano2017deep} to fine-tune the LLM towards better performance and higher data efficiency.

Later in Section~\ref{CV:difficulty}, we will revisit the above procedure and understand it as a natural choice for training an agent to interact with the text environment.

\newpage

\begin{figure*}[!t]
\centering
\includegraphics[width=\linewidth]{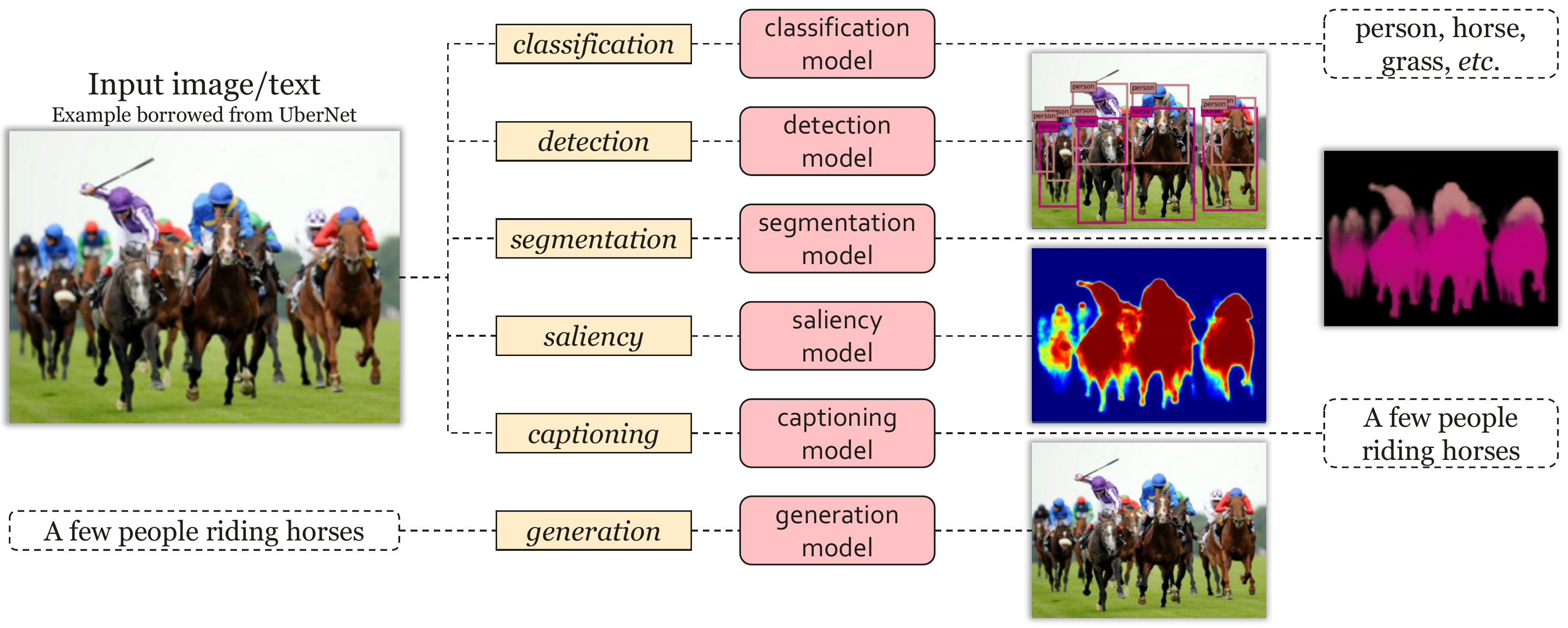}
\caption{The current status of computer vision. Different problems are solved by different models or algorithms. Image credit: UberNet~\cite{kokkinos2017ubernet}.}
\label{fig:CV_status}
\end{figure*}

\section{CV: The Next Battlefield of AGI}
\label{CV}

Humans perceive the world based on multiple data modalities. It is a common knowledge that about 85\% of what we learn is through our vision system. Therefore, given that the NLP community has shown the promise of AGI, it is natural to consider computer vision (CV) or multimodality (which includes at least the vision and language domains) as the next battlefield of AGI.

Here we provide two additional comments to complement the above statement. \textbf{First}, it is clear that CV is a superset of NLP, because humans read articles by first recognizing characters in the captured images and then understanding the contents. In other words, an AGI in CV (or multimodality) should cover all abilities of an AGI in NLP. \textbf{Second}, we argue that language alone is insufficient in many scenarios. For example, when one tries to find detailed information about an unknown object (\textit{e.g.}, animal, fashion, \textit{etc.}), the best way is to capture an image and use it for online search; purely relying on text descriptions can introduce uncertainty and inaccuracy. As another case, as we shall see in Section~\ref{CV:difficulty}, it is not always easy to refer to fine-grained semantics in a scene (for recognition or image editing), and it is more efficient to think in a vision-friendly manner, \textit{e.g.}, using a point or box to locate an object rather than saying something like `the person who is wearing black jacket, standing in front of the yellow car, and talking to another person'.

\subsection{Ideal and Reality}
\label{CV:status}

We desire a CV algorithm that can solve generic tasks, possibly by interacting with the environment. Note that the requirement is not limited to recognizing everything or performing dialog based on an image or video clip. It shall be a holistic system that receives generic orders from humans and produces the desired results. But, the current status of CV is quite preliminary. As shown in Figure~\ref{fig:CV_status}, the CV community has been using different modules and even systems for different vision tasks. Below, we list a few of them.
\begin{itemize}
\item \textbf{Image classification} is one of the most fundamental tasks in CV, due to the simplicity of the setting and the cheapness of collecting training data. State-of-the-art image classification algorithms are based on deep neural networks including convolutional networks~\cite{krizhevsky2012imagenet,he2016deep} and vision transformers~\cite{dosovitskiy2021image,liu2021swin}. A pre-training stage with either self-supervised representation learning~\cite{bao2022beit,he2022masked} or large-scale datasets (\textit{e.g.}, the full ImageNet~\cite{deng2009imagenet} or even external datasets~\cite{sun2017revisiting,yu2022coca}) is very helpful to improve the classification accuracy.
\item The models for \textbf{object detection and instance segmentation} are mostly fine-tuned from the models trained for image classification. Researchers designed specific modules (often referred to as the head) to use the image features extracted by the classification network (often referred to as the backbone) for object localization and recognition. The head modules can be roughly categorized into the two-stage~\cite{girshick2015fast,ren2015faster,he2017mask} and one-stage~\cite{redmon2016you,liu2016ssd} methods, and the transformer blocks have been used~\cite{carion2020end} and pushed the performance on real-world data~\cite{lin2014microsoft} towards a higher level~\cite{zhang2023dino,li2023mask}.
\item The \textbf{semantic segmentation} algorithms fine-tune models trained for image classification in another way. The early efforts involve the encoder-decoder architecture which first downsamples the original image to extract semantic features and then upsamples the features to the original resolution~\cite{long2015fully,chen2017deeplab}. The idea was also inherited to medical images~\cite{ronneberger2015u} and generalized to 3D data~\cite{milletari2016v}. It was shown that keeping high-resolution features improves the segmentation accuracy~\cite{wang2020deep}. Vision transformers also offered new opportunities for more accurate segmentation models~\cite{cheng2021per,xie2021segformer}, especially for more challenging datasets~\cite{zhou2017scene}.
\item The \textbf{image captioning} task~\cite{lin2014microsoft,krishna2017visual} is one of the early trials for cross-modal understanding. In the beginning, pre-trained vision models are equipped with a recurrent head such as LSTM~\cite{hochreiter1997long} for generating captions~\cite{vinyals2015show,xu2015show}. Recently, researchers developed an alternative solution for image captioning which involves fine-tuning foundation models that have connected vision to language~\cite{zhou2020unified,alayrac2022flamingo,li2022blip}.
\item For \textbf{text-to-image generation}, state-of-the-art algorithms~\cite{ramesh2022hierarchical,rombach2022high} are based on the alignment between vision and language. For this purpose, a cross-modal pre-trained model such as CLIP~\cite{radford2021learning} is inherited, based on which probabilistic models are used to decode sequential tokens into images~\cite{razavi2019generating,ramesh2021zero,ding2021cogview} or denoising latent diffusion models~\cite{ramesh2022hierarchical,saharia2022photorealistic,rombach2022high}.
\end{itemize}
Besides, there exist algorithms for other vision tasks, including multiple object tracking~\cite{yan2021learning,meinhardt2022trackformer,zhang2022bytetrack}, pose estimation~\cite{xu2022vitpose}, and many others. It is clear that the current status of CV (individual algorithms are used for different purposes) is far from what the GPT series has achieved in the NLP field.

\subsection{Unification Is the Trend}
\label{CV:unification}

Below, we summarize recent research topics towards unification in CV into five categories.

\begin{figure}[!t]
\centering
\includegraphics[width=\linewidth]{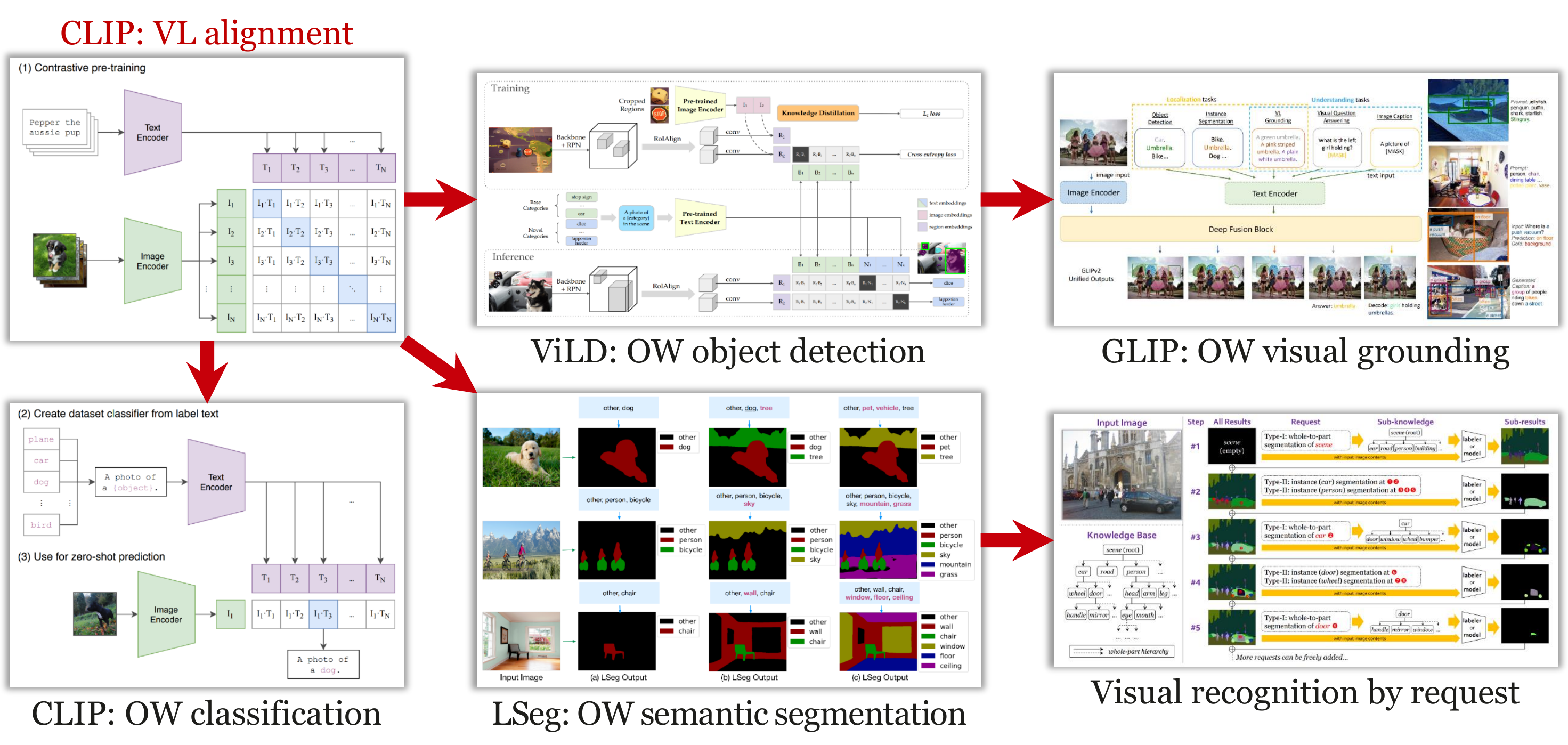}
\caption{Vision-language (VL) pre-training enables open-world (OW) recognition in many vision tasks. Image credit: CLIP~\cite{radford2021learning}, ViLD~\cite{gu2022open}, LSeg~\cite{li2022language}, GLIP~\cite{zhang2022glipv2}, ViRReq~\cite{tang2023visual}.}
\label{fig:CV_openworld}
\end{figure}

\subsubsection{Open-world Visual Recognition}
\label{CV:unification:openworld}

In a long period of time, most CV algorithms can only recognize the concepts that appear in the training data, leading to a `closed-world' of visual concepts. In opposite, the concept of `open-world' refers to the ability that a CV algorithm can recognize or understand any concept regardless whether it has appeared before. The open-world ability\footnote{Sometimes, `open-world' is referred to as `open-set' or `open-domain', although these terminologies may have slightly different meanings.} is often introduced by natural language since it is a natural way for humans to understand new concepts. This explains why language-related tasks such as image captioning~\cite{vinyals2015show,xu2015show} and visual question answering~\cite{malinowski2014multi,malinowski2015ask,andreas2016neural} contributed to the earliest open-world settings for visual recognition.

Recently, with the emergence of vision-language pre-training (\textit{e.g.}, CLIP~\cite{radford2021learning} and ALIGN~\cite{jia2021scaling}), it becomes much easier to align features in the vision and language domains. The unified feature space not only offers simpler pipelines for image captioning~\cite{zhou2020unified,alayrac2022flamingo,li2022blip} and visual question answering~\cite{alayrac2022flamingo,li2023blip,huang2023language}, but also creates a new methodology~\cite{radford2021learning} for conventional visual recognition tasks. For example, image classification can be done by simply matching the query image with a set of templates (also known as `prompts') saying \textsf{a photo of \{something\}}, where \textsf{something} can be any (hence open-world) concept like \textsf{cat} or \textsf{Siberian husky}, and set the result to be the candidate with the highest matching score. Beyond the vanilla version, researchers developed algorithms~\cite{zhou2022learning,wang2022learning} named `learning to prompt' to improve the classification accuracy. Later, the methodology was inherited from image classification to object detection~\cite{gu2022open,du2022learning}, semantic segmentation~\cite{xu2022simple,li2022language}, instance segmentation~\cite{huynh2022open}, panoptic segmentation~\cite{ding2022open,jain2023oneformer}, and further extended to visual grounding~\cite{li2022grounded} and composite visual recognition~\cite{tang2023visual} tasks. These tasks can benefit from vision-language models pre-trained with enhanced localization~\cite{li2022grounded,zhong2022regionclip}.

Open-world visual recognition is closely related to zero-shot visual recognition because both of them try to generalize the recognition ability to the concepts that have not appeared in the training set. However, in the authors' opinion, it is yet unclear whether and how deep learning algorithms can recognize unseen concepts. Indeed, there are some special cases that zero-shot recognition can be achieved (\textit{e.g.}, the training data contains \textsf{dog}, \textsf{cat}, and \textsf{dog's head}, but it does not contain \textsf{cat's head}; it is possible that the algorithm can learn the concept of \textsf{cat's head} from composition without training data), but in most cases, the zero-shot ability was inherited from the pre-trained vision language model. Note that the original CLIP model and other variants (\textit{e.g.}, OpenCLIP~\cite{cherti2023reproducible} and EVA-CLIP~\cite{sun2023eva}) were pre-trained on large-scale image-text pairs which may have contained the target concepts withheld from the downstream training set. Therefore, we argue that `open-world' is a more precise description than `zero-shot'.

As language introduces flexibility to visual recognition, it also brings the drawback of referring to detailed semantics in complex scenes. For example, when a large number of same-class objects appear in an image, it is difficult for the model to ask about the position, shape, or attributes of a specific object. This issue is easily solved in vision itself, \textit{e.g.}, one can use a point to indicate the object of interest. We will get back to this issue in the part discussing multimodal dialog (see Section~\ref{CV:unification:dialog}).

\begin{figure}[!t]
\centering
\includegraphics[width=\linewidth]{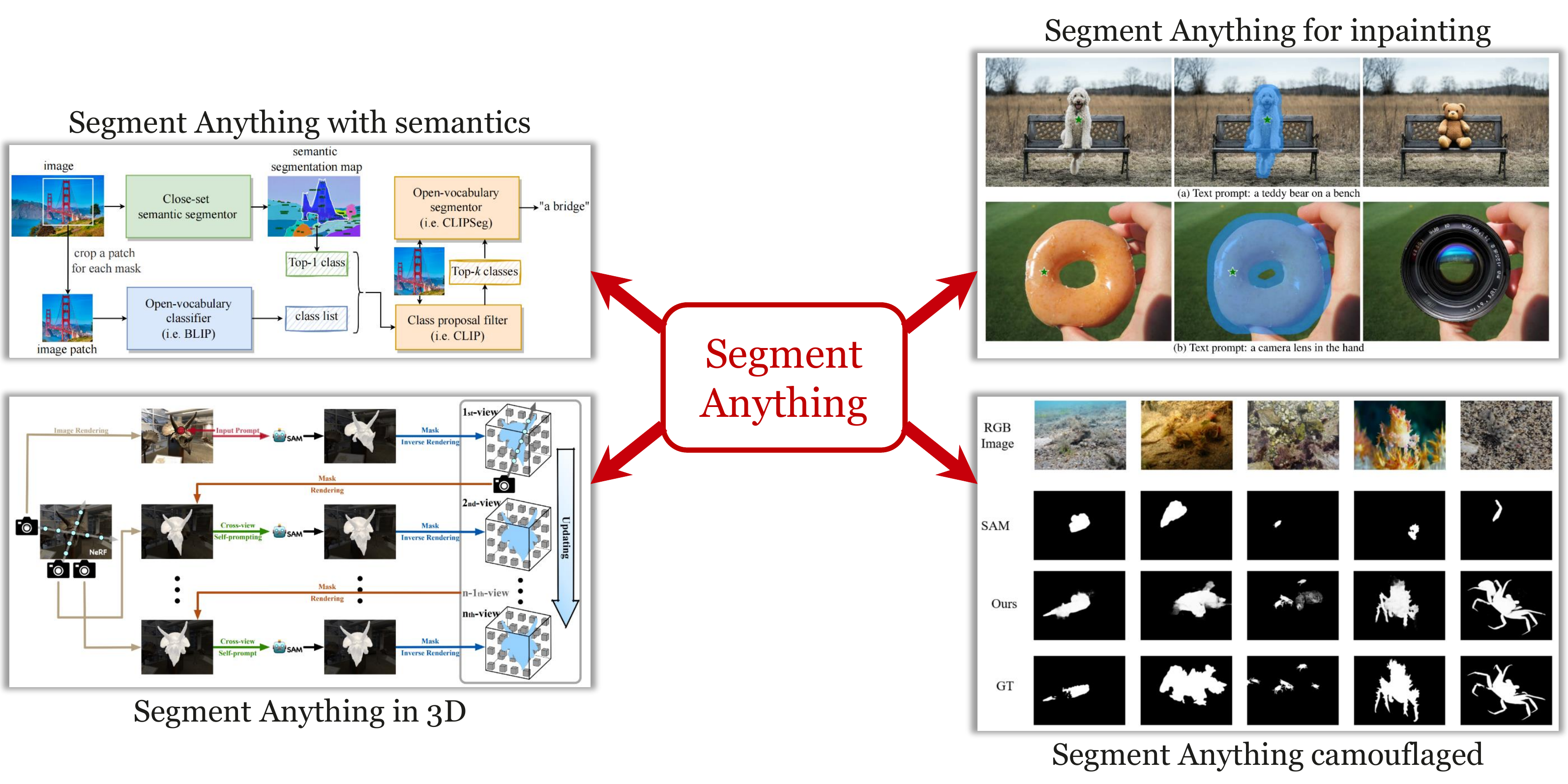}
\caption{The pre-trained SAM~\cite{kirillov2023segment} is easily transferred for various downstream vision tasks. Image credit: SSA~\cite{chen2023semantic}, SA3D~\cite{cen2023segment}, inpaint anything~\cite{yu2023inpaint}, SAM-Adapter~\cite{chen2023sam}.}
\label{fig:CV_anything}
\end{figure}

\subsubsection{The \textit{Segment Anything} Task}
\label{CV:unification:anything}

The \textit{Segment Anything} task~\cite{kirillov2023segment} was introduced recently as a generalized module to cluster raw image pixels into groups, many of which correspond to basic visual units in the image. The proposed task supports several types of prompts including point, contour, text, \textit{etc.}, and produces a few masks as well as scores for each prompt or each combination of prompts. Trained on a large-scale dataset with about $10$ million images, the derived model, SAM, was able to transfer to a wide range of segmentation tasks including medical image analysis~\cite{ma2023segment,deng2023segment,he2023accuracy}, camouflaged object segmentation~\cite{tang2023can,chen2023sam}, 3D object segmentation~\cite{cen2023segment}, object tracking~\cite{yang2023track}, as well as application scenarios such as image inpainting~\cite{yu2023inpaint}. SAM can also be used with state-of-the-art visual recognition algorithms, such as refining bounding boxes produced by visual grounding~\cite{liu2023grounding} algorithms into masks, and feeding the segmented units into open-set classification algorithms for image tagging~\cite{chen2023semantic,zou2023segment}.

Technically, the keys of SAM lie in the prompting mechanism and data closure, \textit{i.e.}, closing the segmentation task with a small amount of feedback from labelers. The unified form of prompts makes SAM look like a part of the vision foundation model or pipeline, but there are still many unsolved issues. For example, it remains unclear about the upstream and downstream modules of SAM (if SAM is indeed part of the pipeline), and SAM can be severely impacted by pixel-level appearance, \textit{e.g.}, an \textsf{arm} can be segmented from the \textsf{torso} exactly on the border of clothes, implying that color is the dominant factor for segmentation. In general, it is likely that SAM has over-fitted to the \textit{Segment Anything} task itself and hence weakened its ability of classification.

\begin{figure}[!t]
\centering
\includegraphics[width=\linewidth]{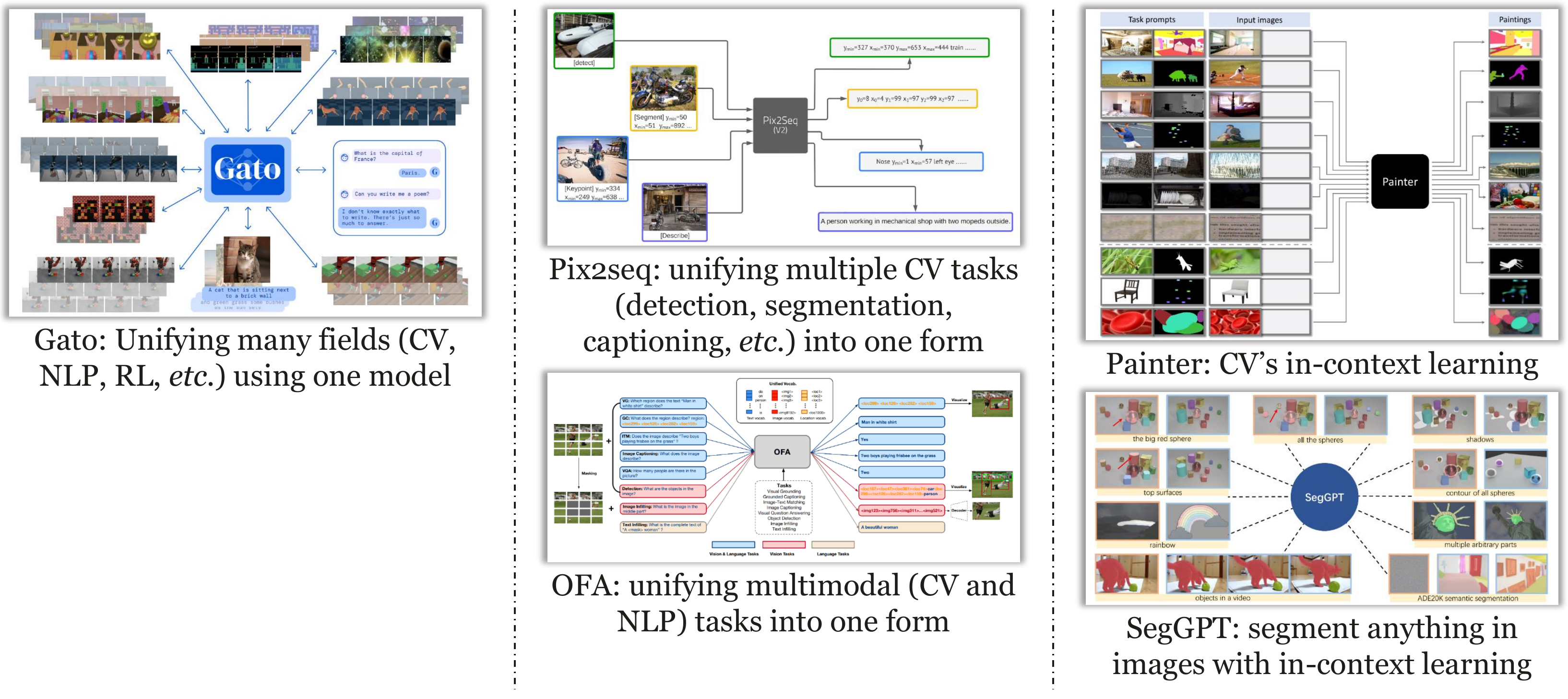}
\caption{Generalized visual encoding allows one model to be trained for various visual and/or multimodal understanding tasks. Image credit: Gato~\cite{reed2022generalist}, pix2seq-v2~\cite{chen2022unified}, OFA~\cite{wang2022ofa}, Painter~\cite{wang2023images}, SegGPT~\cite{wang2023seggpt}.}
\label{fig:CV_encoding}
\end{figure}

\subsubsection{Generalized Visual Encoding}
\label{CV:unification:encoding}

Another way to unify CV tasks is to provide a generalized visual encoding for them. There are several methodologies to achieve this goal.

A key difficulty lies in the large variance between vision tasks, \textit{e.g.}, object detection requires a set of bounding boxes while semantic segmentation requires a dense prediction over the entire image, both of which are very different from a single label required by image classification. As all can understand it, natural language offers a unified form to represent everything. An early effort named pix2seq~\cite{chen2022pix2seq} showed that object detection results (\textit{i.e.}, bounding boxes) can be formulated into natural language and coordinates and then converted into tokens as the output of vision models. In a later version, pix2seq-v2, they generalized the representation to unify the output of object detection, instance segmentation, keypoint detection, and image captioning. Similar ideas were also used for other image recognition~\cite{kolesnikov2022uvim}, video recognition~\cite{yang2023vid2seq}, and multimodal understanding~\cite{lu2022unified,wang2022ofa,zhu2022uni} tasks.

Besides using language, researchers also tried to use vision alone to unify everything. The idea was named in-context learning and was borrowed from the NLP community~\cite{brown2020language}, suggesting that a pre-trained model can realize the intention of new tasks from a few demonstrations. This learning paradigm was first introduced into CV using natural language as prompts~\cite{alayrac2022flamingo}. In~\cite{wang2023images}, different vision tasks, including instance segmentation, keypoint detection, depth estimation, saliency detection, \textit{etc.}, were formulated into assigning different color patches or regions in the output image canvas, hence a single model named \textit{Painter} can be trained to deal with them all. The framework was then extended into a more generalized form which also supports video segmentation~\cite{wang2023seggpt}.

In the backbone of the above algorithms lies the vision transformer~\cite{dosovitskiy2021image}, which offers strong data fitting ability in different modalities. The ability was verified by an earlier work which trained a generalist agent named \textit{Gato}~\cite{reed2022generalist} to unify vision, language, and robotics tasks as long as the desired output can be encoded into a sequence of tokens.

Despite the ability of unified representation, it is questionable how far the methodology has gone beyond multi-task visual representation learning, where different tasks are integrated by incorporating multiple loss functions~\cite{kokkinos2017ubernet}. Recall that GPT applied in-context learning to unify NLP tasks, but CV does not necessarily follow the same direction: this is because CV tasks are mostly discrete (\textit{e.g.}, there is no intermediate task between segmentation and tracking) and thus there might not be a significant difference between individual and joint optimization strategies.

\begin{figure}[!t]
\centering
\includegraphics[width=\linewidth]{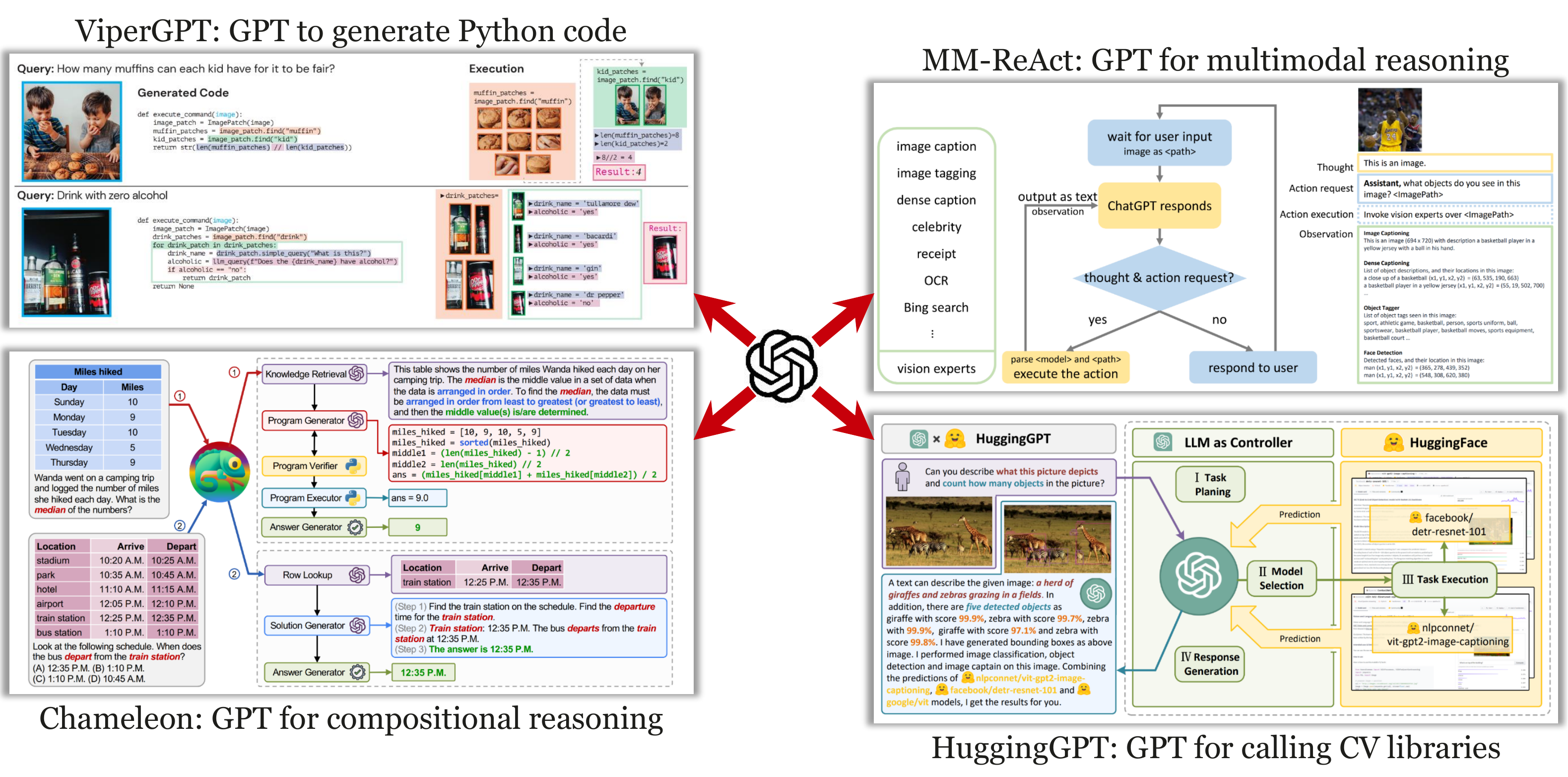}
\caption{GPT offers an easy and unified way to generate code or explanations for visual understanding. Image credit: ViperGPT~\cite{suris2023vipergpt}, Chameleon~\cite{lu2023chameleon}, MM-ReAct~\cite{yang2023mm}, HuggingGPT~\cite{shen2023hugginggpt}.}
\label{fig:CV_understanding}
\end{figure}

\subsubsection{LLM-guided Visual Understanding}
\label{CV:unification:understanding}

Visual recognition can be complex especially when it involves compositional concepts and/or relationships between visual instances. It is difficult for end-to-end models (vision-language pre-trained models for visual question answering~\cite{alayrac2022flamingo,li2023blip,huang2023language}) to produce answers following a procedure that is easily understood by humans.

To alleviate the issue, a practical methodology lies in generating explainable logic to assist visual recognition. The idea was not new. Several years ago, prior to the appearance of the transformer architecture, researchers proposed to use the long short-term memory (LSTM) model~\cite{hochreiter1997long} to generate programs so that vision modules are invoked as modules for complex question answering~\cite{johnson2017inferring}. At that time, the ability of LSTM largely limits the idea within the range of relatively simple and templated questions.

Recently, the appearance of large language models (especially the GPT series) makes the conversion of arbitrary questions possible. Specifically, GPT can interact with humans in different ways. For example, it can summarize basic recognition results to the final answer~\cite{yang2023mm} or generate code~\cite{suris2023vipergpt,lu2023chameleon} or natural language scripts~\cite{shen2023hugginggpt} to call basic vision modules. As a result, visual questions can be decomposed into basic modules. This is especially effective for logical questions, \textit{e.g.}, that asking about the spatial relationship between objects or that depending on the number of objects.

LLMs may understand the logic, but they have not yet showed the ability to assist fundamental visual recognition modules. That said, the answer will still be wrong once the basic recognition results are incorrect, \textit{e.g.}, the detection algorithm misses a few objects that are small and/or partially occluded. We expect an essential visual logic to be formulated in the future (\textit{e.g.}, the algorithm can follow a sequential algorithm to detect every object, or be guided by commonsense~\cite{zellers2019recognition} to solve hard cases), possibly with the assistance of LLMs, so that fundamental visual recognition is boosted.

\begin{figure}[!t]
\centering
\includegraphics[width=\linewidth]{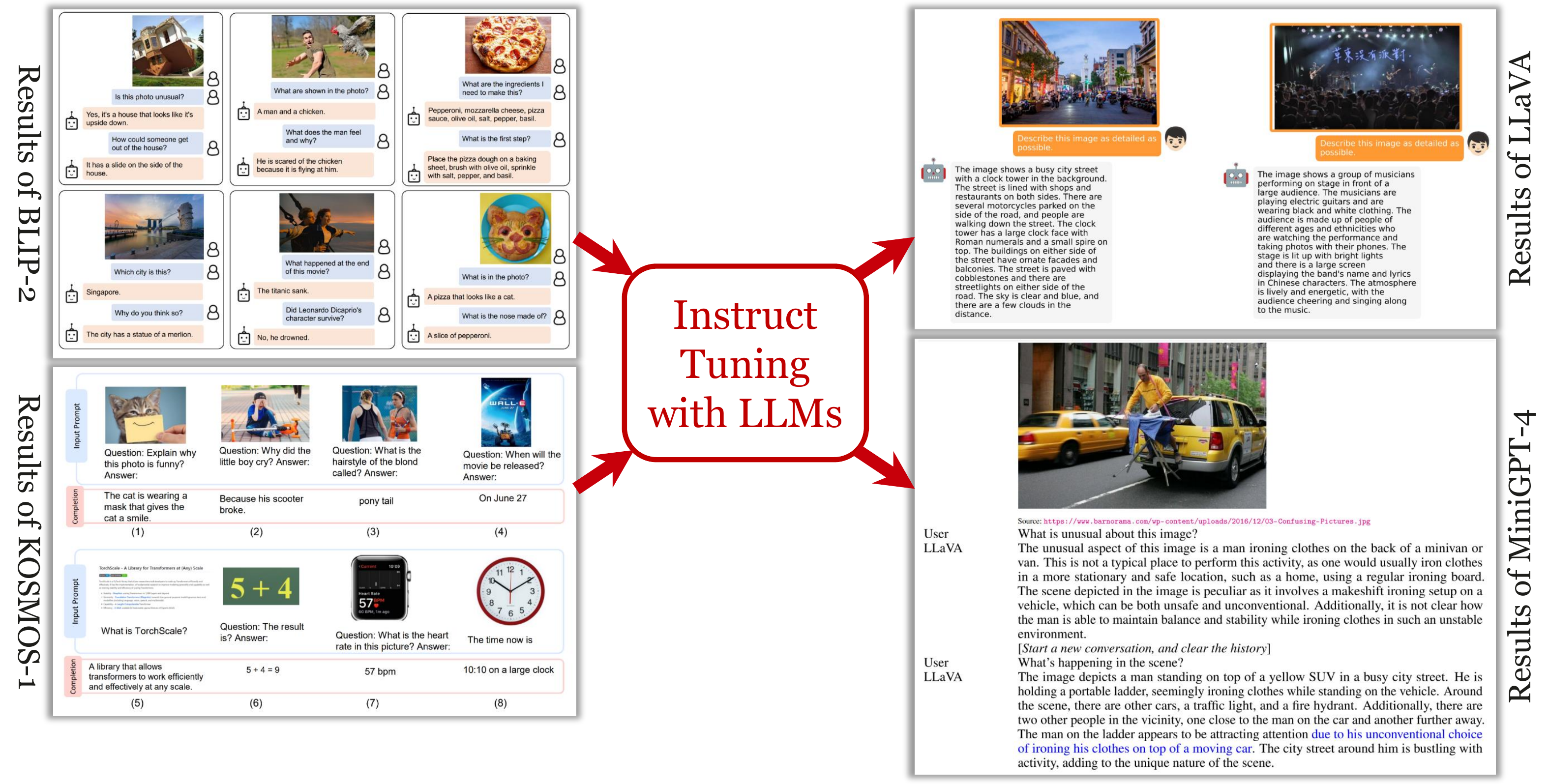}
\caption{With instruction tuning, visual question answering algorithms are extended to multimodal dialog systems. Image credit: BLIP-2~\cite{li2023blip}, KOSMOS-1~\cite{huang2023language}, LLaVA~\cite{liu2023visual}, MiniGPT-4~\cite{zhu2023minigpt}.}
\label{fig:CV_understanding}
\end{figure}

\subsubsection{Multimodal Dialog}
\label{CV:unification:dialog}

Multimodal dialog extends text-based dialog to the vision domain. The early efforts involved visual question answering in which various datasets with simple questions have been constructed~\cite{antol2015vqa,goyal2017making,johnson2017clevr}. With the rapid development of LLMs, multi-round question answering was made available by fine-tuning pre-trained vision and language models together~\cite{li2023blip,huang2023language}\footnote{GPT-4~\cite{openai2023gpt} also showed examples of multimodal dialog, but it is unclear how they achieved the ability, especially for the cases with rich texts, \textit{e.g.}, solving a complex physical problem and understanding a joke which is mainly described in optical characters.}. It was also shown that a wide range of questions can be answered via in-context learning in multimodality~\cite{alayrac2022flamingo} or using GPT as the logic controller~\cite{wu2023visual}.

Recently, a novel paradigm developed in the GPT series, named instruct learning~\cite{ouyang2022training}, has been inherited to enhance the quality of multimodal dialog~\cite{liu2023visual,zhu2023minigpt}. The idea was to provide a few reference data (\textit{e.g.}, objects, descriptions) from ground-truth annotation or recognition results and ask the GPT model to generate instruction data (\textit{i.e.}, enriched question-answer pairs). Fine-tuned with these data (without reference), the foundation models for vision and language can interact with each other via a lightweight network module (\textit{e.g.}, a Q-former~\cite{li2023blip}).

Multimodal dialog offers a preliminary interactive benchmark for computer vision, but, as a language-guided task, it also has the weaknesses analyzed in the open-world visual recognition (see Section~\ref{CV:unification:openworld}). We expect that enriching the form of queries (\textit{e.g.}, using generalized visual encoding methods, see Section~\ref{CV:unification:encoding}) can push multimodal dialog to a higher level.

\subsection{The Essential Difficulty}
\label{CV:difficulty}

Indeed, the above efforts have largely advanced the progress of unification in CV. However, the community is still far from an algorithm that can solve a wide range of real-world tasks, in particular when interaction is needed. In this part, we discuss on the essential difficulty that leads to the current dilemma.

\subsubsection{GPT Revisited}
\label{CV:difficulty:revisited}

We start with recalling the definition of AGI (see Section~\ref{AGI}, the definition was inherited from~\cite{goertzel2007artificial,silver2021reward}). In short, the goal of AGI is to maximize a reward in an interactable environment.

GPT defined such an environment with the chat task. Note that, in a pure-text world, chat is the perfect task for an agent to learn from interaction (talking with humans and receiving feedback); meanwhile, any task can be defined by chat. In our opinion, establishing the environment (with the chat task) is the most important insight of GPT. The technical solutions (\textit{i.e.}, generative pre-training followed by instruct fine-tuning) can be derived from the chat task: generative pre-training is to memorize the distribution of the environment (world); instruct fine-tuning is to align the learned contribution with question-answer pairs for problem-solving. There are clear boundary between them, as the fine-tuning stage is driven by specific tasks while the pre-training stage is not.

We try to build the relationship between the basic elements of an environment and GPT. We find that the state, action, and reward in the environment correspond to the prompting mechanism, the desired target, and the feedback from users, respectively. We expect that CV algorithms are also trained in such an environment, or at least with the above factors clearly defined.

\subsubsection{Why Not Establishing Environments for CV?}
\label{CV:difficulty:environments}

Conceptually, an AGI in CV should also be trained in environments. Back to the 1970s, David Marr pointed out that CV algorithms should construct a world model and learn abilities by interacting with the model~\cite{marr1982vision}. Other pioneers in AI, including Hans Moravec and Rodney Brooks, also emphasized the importance of learning from environments~\cite{moravec1988mind,brooks1990elephants}. However, establishing environments for CV is never an easy task, unlike that for NLP which is quite straightforward.

There are mainly two options, namely, training agents in the real world or in virtual, simulated environments. The former option is closer to the final objective, but the over-high costs and uncontrollable safety issues have constrained it in small-scale and toy scenarios (\textit{e.g.}, training robotic arms for object grasping). The latter option is relatively easy to achieve, but it suffers the fidelity issues (not only about 3D modeling and rendering, but also about the behavior of other agents) and thus has to alleviate a significant domain gap when being transferred to the real world.

Due to the high difficulty of simulating the world (\textit{i.e.}, establishing environments), the CV community has taken an alternative solution which is to sample the world. It involves two major steps, namely, image/video data collection and semantic annotation. The first step is to perform a sparse sampling of the real world -- note that, although their size has significantly increased during the past decades, all existing datasets are still orders of magnitude smaller than the real world. The second step is to expect what agents need in order to accomplish tasks and convert the requirements (\textit{e.g.}, detecting objects) into semantic annotations. From this point of view, we refer to them \textbf{proxy tasks} as they serve as surrogates to achieve the goal of AGI. Note that proxy tasks exist in almost all AI fields, \textit{e.g.}, in NLP, there are various such tasks including translation, named entity recognition, and others.

\begin{figure}[!t]
\centering
\includegraphics[width=\linewidth]{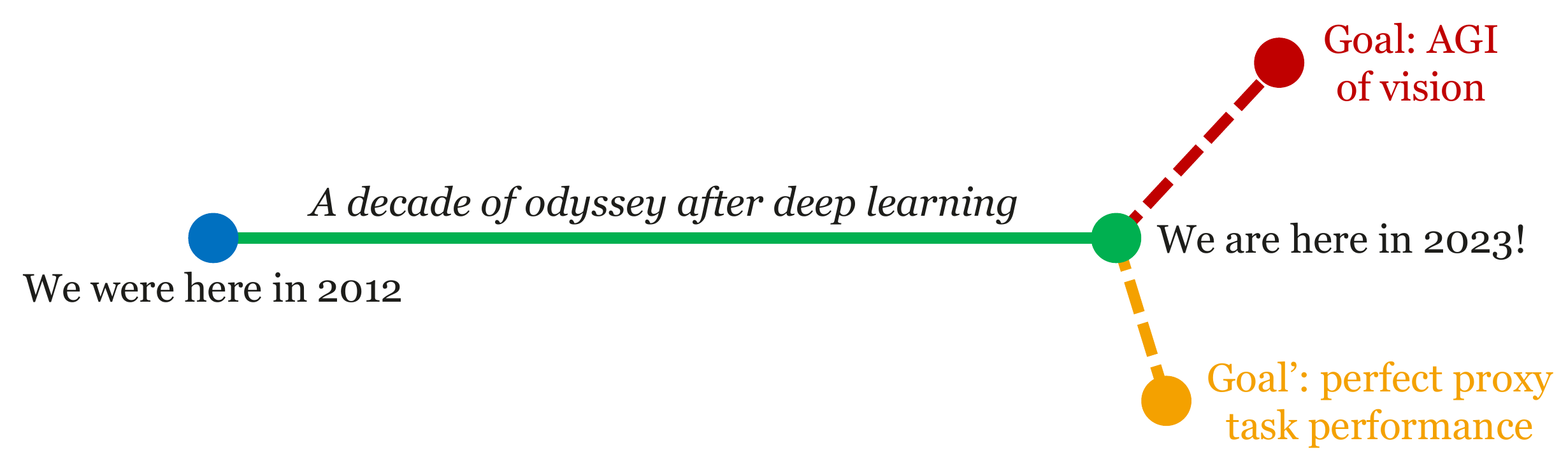}
\caption{An illustration of our opinion, \textit{i.e.}, the proxy tasks in CV are dying.}
\label{fig:CV_proxy}
\end{figure}

\subsubsection{Proxy Is Dying!}
\label{CV:difficulty:proxy}

The proxy tasks have been vastly improved in the past decade, thanks to the rapid development of deep learning. One of the most well-known examples lies in ImageNet-1K classification~\cite{deng2009imagenet,russakovsky2015imagenet}, where the best accuracy is under $50\%$ prior to AlexNet~\cite{krizhevsky2012imagenet}, while the accuracy is higher than $90\%$ as of today~\cite{yu2022coca}. The odyssey was made possible by strong network architectures, effective optimization tricks, external training data, \textit{etc}. Nevertheless, many research papers have been still pursuing higher accuracy on this dataset. Standing upon a high baseline (\textit{e.g.}, $85\%$), the improvement brought by the proposed algorithm is often small (\textit{e.g.}, $0.5\%$), leading to a weird situation in that implementation tricks contribute even more than the proposed algorithm itself.

We illustrate the situation in Figure~\ref{fig:CV_proxy}. Let us assume that AGI of vision and perfect performance of proxy tasks are two goals in the space of algorithms. In the pre-deep learning era, CV algorithms are mostly weak, so setting the goal to be high proxy task performance is reasonable. As of today, CV algorithms have been much stronger than before; consequently, continuing to improve proxy tasks can drive us away from AGI. We refer the readers to what happened in NLP: GPT offered a unified solution and largely reduced the importance of the conventional proxy tasks (\textit{e.g.}, translation, named entity recognition, \textit{etc.})\footnote{Disclaimer: these tasks are still important in some real-world applications, but it is unlikely that they shall be studied in an old-fashioned manner. With the chat task, these tasks can be accomplished with simple prompts.}.

\section{Future: Learning from Environment}
\label{future}

The above analysis calls for a new paradigm for training strong agents for CV. In this section, we convert our opinions and insights into an imaginary pipeline, review existing works that are related to the pipeline, and make comments on future research directions based on the pipeline.

\begin{figure*}[!t]
\centering
\includegraphics[width=\linewidth]{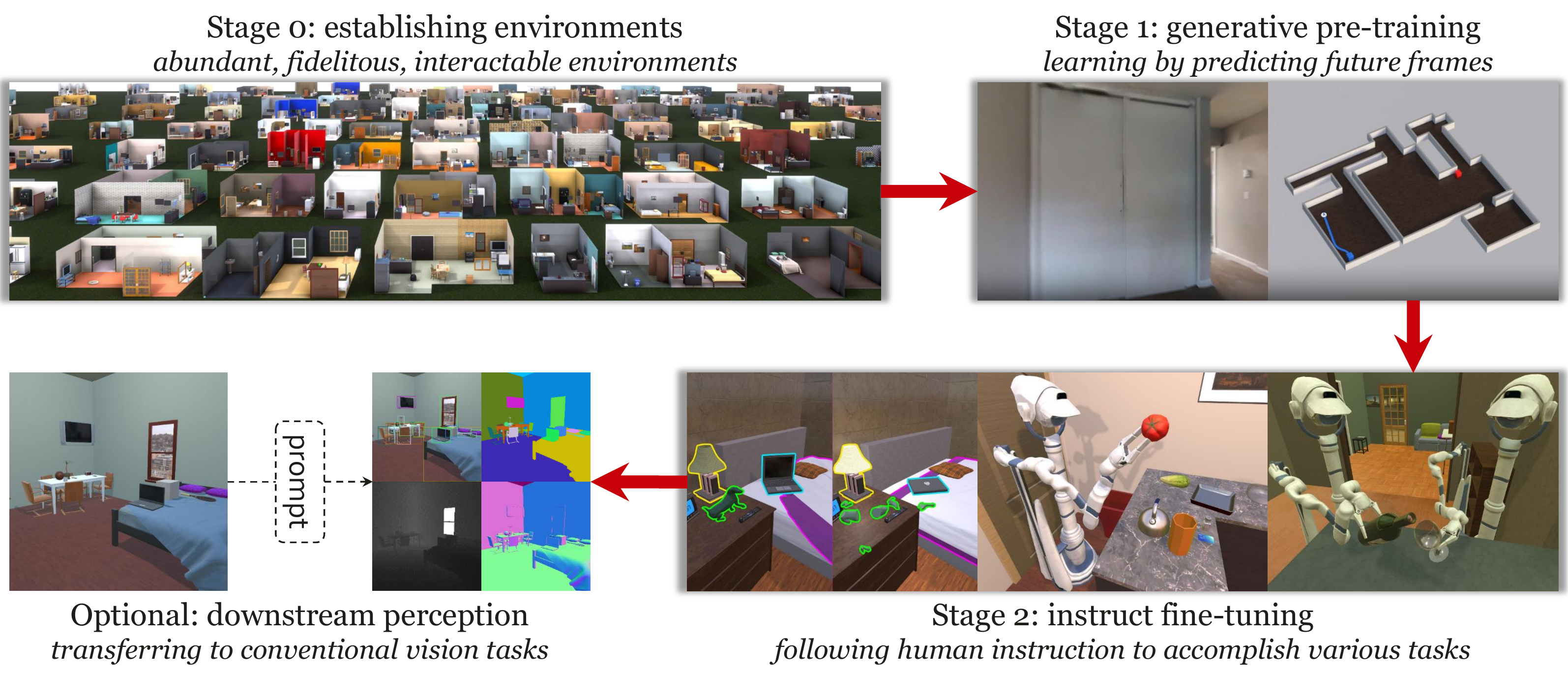}
\caption{An imaginary pipeline of training a stronger agent for computer vision. The idea was borrowed from GPT, with Stages 1 and 2 performing generative pre-training and instruct fine-tuning, respectively. Differently, environments need to be established prior to the pre-training stage, which itself is a major challenge. Image credit: the Stage 1 image is from Habitat~\cite{savva2019habitat} and others are from ProcTHOR~\cite{deitke2022procthor}.}
\label{fig:future_pipeline}
\end{figure*}

\subsection{An Imaginary Pipeline}
\label{future:pipeline}

Figure~\ref{fig:future_pipeline} shows our idea. The pipeline comprises three stages: Stage 0 for establishing environments, Stage 1 for pre-training, and Stage 2 for fine-tuning. When necessary, the fine-tuned model can be prompted for conventional visual recognition tasks. Below, we describe each stage in detail.
\begin{itemize}
\item\textbf{Stage 0: establishing environments.} As analyzed previously, high-quality environments are strongly required for AGI in CV. Here the concept of `high-quality' includes but is not limited to abundance (there should be sufficient and diversified environments), fidelity (visual appearance and other agents' behavior should be close to the real world), and richness in interaction (the agent can be asked to perform a wide range of tasks by interacting with the environments).
\item\textbf{Stage 1: generative pre-training.} The algorithm is asked to explore the environments and pre-trained to predict future frames. The biggest difference from the GPT task (predicting the next token) in NLP lies in that the future frames depend on the action of the agent (in NLP, the pre-trained text corpus remains unchanged), so the model is trying to learn a joint distribution of state and action. This strategy is especially useful when the set of established environments is insufficient to approximate the world's distribution. Note that, as CV is a superset of NLP (see the paragraph before Section~\ref{CV:status}), the size (\textit{e.g.}, number of parameters) of pre-trained CV models should be orders of magnitude larger than NLP models.
\item\textbf{Stage 2: instruct fine-tuning.} The pre-trained model is guided to accomplish real-world tasks, following human instructions. Intuitively, there are much more types of allowed interaction between the agent and environments, including exploration, navigation, using language, performing physical actions, and many others. A reasonable conjecture is that much more instruction data should be collected, which also corresponds to the size of foundation CV models.
\item\textbf{Optional: downstream perception.} We expect that CV algorithms can learn all required abilities of perception from the previous stage, \textit{e.g.}, in order to accomplish a very simple task, \textit{Buy me a cup of coffee}, the model must at least learn to (i) explore around with safety, (ii) recognize where the coffee bar is, (iii) communicate with the shop assistant with language, and (iv) grasp the bought coffee. Such a model, when properly prompted, should output desired perception results, including tracking another agent (for not colliding with it),  open-set visual recognition (for finding the bar and the bought coffee), and others. This is related to the idea of analysis by synthesis~\cite{yuille2006vision}.
\end{itemize}

To sum up, we expect the agent to be task-oriented, \textit{i.e.}, focusing itself on accomplishing tasks in the established environments. The proxy tasks (see Section~\ref{CV:difficulty:proxy}) are to be solved naturally with prompts.

\subsection{Existing Works}
\label{future:relatedworks}

We briefly review the existing works that are related to the imaginary pipeline.

\subsubsection{3D Embodied Environments}
\label{future:relatedworks:environments}

There are typically two options for establishing virtual environments. The first option is to collect visual content from a real-world scenario and perform 3D reconstruction. For example, Habitat~\cite{savva2019habitat} released more than $200$ scanned scenarios for visual navigation. The second option is to generate (render) scenes with 3D models. For example, ProcTHOR~\cite{deitke2022procthor} provided a large set of 3D objects and a program to randomly generate room layouts, so that one can sample an arbitrary number of 3D scenes and perform various tasks including navigation, grasping, and reordering.

However, the existing environments (including the above two and many others~\cite{zhu2017target,kolve2017ai2}) have not yet validated the ability to scale up to the level of the real world. In particular, when more and more environments are sampled from ProcTHOR~\cite{deitke2022procthor}, the performance, either for embodied or downstream recognition tasks, can saturate quickly. Clearly, for any existing method, there is a tradeoff between abundance and fidelity. This is a major challenge before pre-trained CV models can exhibit the scaling law~\cite{kaplan2020scaling} and emergent abilities~\cite{wei2022emergent} as NLP models did.

\subsubsection{Pre-training Vision Models}
\label{future:relatedworks:pretraining}

In the past years, visual pre-training methods have been largely developed. Contrastive learning (CL)~\cite{wu2018unsupervised,oord2018representation,chen2020simple,he2020momentum} offered the first methodology to surpass supervised learning in downstream tasks, and masked image modeling (MIM)~\cite{bao2022beit,he2022masked,xie2022simmim} pushed the performance of pre-trained models to a higher level. The major difference between them lies in the pre-training objective, where CL is discriminative and MIM is generative.

Built upon a generative objective, MIM is closer to what we desire in the aforementioned pipeline. However, predicting future frames in environments is different from predicting missing contents in sampled images or videos, which is similar to the difference between masked language modeling~\cite{devlin2019bert} and generative pre-training~\cite{radford2018improving}. Additionally, compared to text data, there can be much more redundant information in vision data. We conjecture that data compression is an important factor in the pre-training task.

\subsubsection{Reinforcement Learning for Game Playing}
\label{future:relatedworks:gaming}

Interacting with environments is closely related to game playing. The past decade has witnessed the integration of deep learning and reinforcement learning, resulting in a series of algorithms for playing various games, such as Atari 2600 games~\cite{mnih2015human,schulman2017proximal}, simulated robotics tasks~\cite{schulman2015trust,schulman2017proximal}, Go and other chess~\cite{silver2016mastering,silver2017mastering}, StarCraft II~\cite{vinyals2019grandmaster}, and so on. Effective algorithms were also designed for combining multiple reinforcement learning strategies~\cite{hessel2018rainbow} or completing very complex tasks~\cite{ecoffet2021first}.

Compared to the above problems, the real world is much more complicated and involves actions from different aspects including language and robotics. A practical strategy is to first disable some types of interaction to simplify the tasks and add them back when the foundation models are sufficiently strong.

\subsubsection{Embodied Computer Vision}
\label{future:relatedworks:embodied}

Embodied AI refers to the research field in that agents learn from/for interacting with virtual environments. The motivation partly came from how humans learn as babies~\cite{smith2005development}. In the scope of CV, typical tasks include exploration~\cite{pathak2017curiosity,chen2019learning,chaplot2020semantic} where the goal is simply to explore and reconstruct the world, visual navigation~\cite{zhu2017target,anderson2018evaluation,chaplot2020neural} where the agents are asked to explore the world for specific targets (\textit{e.g.}, an image or an object), and embodied question answering~\cite{das2018embodied,gordon2018iqa} where the agents answer questions based on the interaction with the world.

Recently, two works were brought to our attention. The first one is PaLM-E~\cite{driess2023palm} where a general-purpose vision-language model is trained to perform a wide range of embodied tasks. These tasks are organized into a unified prompting system, thanks to the in introduction of LLMs. The second one is ENTL~\cite{kotar2023entl} where an end-to-end system was designed and different stages in embodied CV (including world modeling, localization, and imitation learning) were tokenized and integrated into a sequence prediction task. Both works pushed the unification in embodied CV forward from different directions.

We emphasize that, despite the existing efforts, the real-world scenario is much more complicated than what we have ever created or simulated (see Section~\ref{future:relatedworks:environments}). To achieve the goal of AGI, more interaction types should be supported, long-range and complex tasks should be designed, and instruction data should be collected. System design and engineering might be more important than one thinks.

\subsection{Comments on Research Directions}
\label{future:directions}

As the final part, we make comments on future research directions. With the major goals migrated from the performance of proxy tasks to learning from environments, many popular research directions may have to adjust their goal. Here is a disclaimer: all the following statements are our personal opinions and may be wrong.

\subsubsection{On Establishing Environments}
\label{future:directions:environments}

A clear goal is to continue increasing the size, diversity, and fidelity of the virtual environments. There are multiple techniques that can help. For example, novel 3D representation forms (\textit{e.g.}, neural rendering field, NeRF~\cite{mildenhall2021nerf,yu2021pixelnerf,zhang2020nerf++}) may be more efficient in achieving a tradeoff between the reconstruction quality and overhead.

Another important direction lies in the richness of environments. It is a non-trivial task to define new, complex tasks and unify them into a prompting system. Also, AI algorithms can benefit much from a better simulation of other agents' behavior~\cite{petrenko2021megaverse} because it can largely improve the abundance of environments and hence the robustness of the trained algorithms.

\subsubsection{On Generative Pre-training}
\label{future:directions:pretraining}

There are mainly two factors that affect the pre-training stage, namely, neural architecture design and proxy task design. The latter is clearly more important and the former shall be built upon the latter.

As analyzed in Section~\ref{future:relatedworks:pretraining}, existing pre-training tasks, including contrastive learning and masked image modeling, shall be modified to allow for efficient explorations in virtual environments. We expect the newly designed proxy to focus on data compression, because there is much heavier redundancy in vision data than in language data. The new pre-training proxy defines the requirement of neural architectures, \textit{e.g.}, to achieve a tradeoff between data compression and visual recognition, the designed architecture should be equipped with an ability to extract different levels (granularity) of visual features by request.

Additionally, cross-modality (\textit{e.g.}, text-to-image) generation will become a direct metric to measure the performance of pre-training. When a unified tokenization method is available, it can be formulated into a multimodal version of the reconstruction loss.

\subsubsection{On Instruct Fine-tuning}
\label{future:directions:environments}

We have not yet entered the scope of defining tasks in the new paradigm. As real-world tasks can be very complicated, we conjecture that some elementary tasks (\textit{e.g.}, exploration, fetching, interaction, \textit{etc.}) can be defined and trained first, so that complex tasks can be decomposed into them. For this purpose, a unified prompting system should be designed and abundant human instructions should be collected. As a reasonable conjecture, the amount of instruction data can be orders of magnitude larger than what has been collected for training GPT and other chatbots.

This is a completely new story for CV. The road ahead is filled with unknown difficulties and uncertainty. We cannot see much at the current point, but clear paths will emerge in the future.

\section{Conclusions}
\label{conclusions}

In this paper, we discuss how to advance CV algorithms towards AGI. We start with reviewing the current status and recent efforts of CV for unification, and then we inherit ideas and insights from NLP, especially the GPT series. Our conclusion is that CV lacks a paradigm that allows learning from environments, for which we propose an imaginary pipeline. We expect that substantial technical evolution is required to bring the pipeline to truth.

\ifCLASSOPTIONcompsoc
  \section*{Acknowledgments}
\else
  \section*{Acknowledgment}
\fi

The authors would like to thank many colleagues and collaborating researchers for instructive discussions.

\bibliographystyle{IEEEtran}
\bibliography{bare_jrnl_compsoc}

\end{document}